\documentclass{bmvc2k}
\usepackage{times}
\usepackage{epsfig}
\usepackage{graphicx}
\usepackage{amsmath}
\usepackage{amssymb}
\usepackage{booktabs}
\usepackage{caption}
\usepackage{hyperref}
\usepackage{booktabs}

\newcommand{\xz}[1]{\textcolor{black}{{#1}}}

\title{Few-Shot Temporal Action Localization  \\ with Query Adaptive Transformer}

\addauthor{Sauradip Nag}{s.nag@surrey.ac.uk}{1,2}
\addauthor{Xiatian Zhu }{xiatian.zhu@surrey.ac.uk}{1}
\addauthor{Tao Xiang}{t.xiang@surrey.ac.uk}{1,2}

\addinstitution{
 Centre for Vision Speech and Signal Processing (CVSSP)\\
 University of Surrey, UK\\
}

\addinstitution{
 iFlyTek-Surrey Joint Research \\
 Centre on Artificial Intelligence
}

\runninghead{Nag, Zhu, Xiang}{Few-Shot TAL with Query Adaptive Transformer}

\def\eg{\emph{e.g}\bmvaOneDot}
\def\ie{\emph{i.e}\bmvaOneDot}

\begin{document}

\maketitle

\begin{abstract}
Existing temporal action localization (TAL) works rely on  a  large  number  of  training  videos  with  exhaustive segment-level annotation, preventing them from scaling to new classes.   As a solution to this problem,  few-shot TAL (FS-TAL) aims to adapt a model to a new class represented by as few as a single video. Exiting FS-TAL methods assume trimmed training videos for new classes. However, this setting is not only unnatural – actions are typically captured in untrimmed videos, but also ignores background video segments containing vital contextual cues for foreground action segmentation. In this work, we first propose a new FS-TAL setting by proposing to use untrimmed training videos. Further,  a  novel  FS-TAL  model  is  proposed  which  maximizes the knowledge transfer from training classes whilst enabling the model to be dynamically adapted to both the new class and each video of that class simultaneously. This is  achieved  by  introducing  a  query  adaptive  Transformer in the model. Extensive experiments on two action localization benchmarks demonstrate that our method can outperform all the state-of-the-art alternatives significantly in both single-domain and cross-domain scenarios. The source code can be found in \href{https://github.com/sauradip/fewshotQAT}{https://github.com/sauradip/fewshotQAT}
\end{abstract}

\section{Introduction}

Temporal action localization (TAL) aims to identify the temporal duration (\ie, the start and end points) and class label of action instances in naturally untrimmed videos \cite{jiang2014thumos,caba2015activitynet}. 
Existing TAL methods \cite{lin2018bsn,xu2020gtad,buch2017sst,wang2017untrimmednets,zhao2017temporal}  use training datasets composed of a large number of videos (e.g., hundreds) per class  with exhaustive segment-level annotation. The annotation is tedious and costly. Further, for some rare classes collecting sufficient video instances may not even be feasible.  
These have severely limited the scalability  and general usability of existing TAL methods.
Inspired by the success of few-shot image classification \cite{sung2018learning,snell2017prototypical,finn2017model,chen2019closer,rusu2018meta}, few-shot learning (FSL) has been recently introduced to TAL \cite{yang2018one,yang2020localizing,zhang2020metal}.
A few-shot learning model is designed to eliminate the annotation of large training data.
This is achieved by meta-learning which enables a model to adapt to any new class with as few as a single video. One of the key challenges in FS-TAL is how to capture the intra-class variation using only a handful (\eg, 1-5) training instances of a new class. One of the key objectives of meta-learning is to thus transfer such intra-class variation information from a large set of seen training classes to the new class to compensate for lack of training data. 

Nonetheless, existing few-shot TAL (FS-TAL) methods 
\cite{yang2018one,yang2020localizing,zhang2020metal}
all adopt a setting under which trimmed videos are used to represent the new classes for model adaptation. This setting seems to problematic:
(1) As mentioned earlier, the TAL problem exists because  most action instances are first captured in untrimmed videos sandwiched by background segments. An analogy is that objects always co-exist with background (\eg, tree/road/wall) in an image. So to obtain the trimmed new class video, one needs to first manually annotate (trim) the untrimmed videos. This begs the question: \textit{why not use the untrimmed video together with the annotation for model adaptation?}
(2) Each new action class occurs in its own specific context (background), which carries important cues on how to segment it. Using trimmed videos means that a FS-TAL model is unable to exploit the contextual information for both knowledge transfer from seen classes and new unseen class adaptation. 

In this work, we first introduce a new and more practical few-shot TAL (FS-TAL) problem setting.
During both the training (meta-learning) and inference (model adaptation) stages, each class is represented by a support set comprising untrimmed videos with temporal annotation. A segmentation model is then built using the support set and applied to a query set containing untrimmed videos of the same class to locate the foreground action instances. This change of setting means that instead of meta-learning a model to temporally align the support set instances with the foreground segments of the query video as in \cite{yang2018one,yang2020localizing,zhang2020metal}, we aim to meta-learn a foreground/background classifier that can be quickly adapted to new classes.
To this end,
we propose a novel FS-TAL model which meta-learns a query adaptive Transformer (QAT) for fast adaption of foreground/background classifier to a new class.
\xz{In particular, this leverages the attention mechanism across the query video and few-shot classifier in order to better capture the intra-class invariance.}
As shown in Figure \ref{fig:network}, our model has two key components, 
a snippet classifier that labels each video snippet into foreground or background, and a query adaptation module designed for query video adaptation.  
The former is a simple binary classifier constructed using the annotated untrimmed support set videos. The latter is formulated as a Transformer that takes both the classifier weight vector and query video features as input and outputs an updated classifier adapted to each query video. This QAT module is meta-learned and fixed during inference; therefore the whole model is inductive. Importantly, our model is flexible 
in that it can work in both the new setting proposed in this paper and the existing setting with trimmed support set.
We make the following contributions:
{\bf (1)}
We introduce a new and more practical FS-TAL problem setting.
{\bf (2)}
We propose a novel FS-TAL model with a query adaptive Transformer for model adaptation to both a given new class and each query video.
{\bf (3)}
Extensive experiments show that the proposed method yields new state-of-the-art performance on two TAL datasets (ActivityNet-v1.3 and Thumos'14).
Under a more challenging and more realistic cross-domain setting, the advantage of our method remains.
\vspace{-0.2in}
\section{Related Works}

{\bf Temporal Action Localization} 
%
An intuitive approach to temporal action localization (TAL) is based on sliding window --
first generating multi-scale segments and then classifying them
\cite{dai2017temporal}.
A key limitation with this pipeline is that 
a large number (thousands) of possible segments are necessary 
for achieving reasonable accuracy, which is computationally expensive.
To overcome this issue, foreground/background modeling is introduced
to generate action proposals  \cite{shou2016temporal,shou2016temporal,gao2017turn,zhao2017temporal,lin2018bsn,lin2019bmn}.
When proposal generation is poor,
incorporating sliding windows could be helpful
\cite{gao2018ctap}.
For improving local segment-level feature representation, \cite{zeng2019graph,xu2020gtad} exploit graph convolutional networks to capture long-range contextual information.
Nonetheless, assuming a pre-collected dataset of all action class during training, all these methods have poor scalability to large number of classes, due to the high annotation cost.

\noindent{\bf Few-shot Learning}
For fast adaptation of a model to any given new class with few training samples,
few-shot learning (FSL) provides a solution \cite{vinyals2016matching, sung2018learning, snell2017prototypical}.
It is often realized by meta-learning which simulates the behaviour of new tasks with novel classes represented by only a handful of labeled samples.
This eliminates the requirement of labeling a large dataset
for a new class.
Representative approaches include
hallucination (data augmentation) \cite{hariharan2017low, wang2018low},
initialization optimization \cite{finn2017model,rusu2018meta,ravi2016optimization},
metric learning \cite{koch2015siamese,sung2018learning}.
Beyond image classification, FSL has also been introduced to object detection \cite{kang2019few,dong2018few,hu2019silco} and semantic segmentation \cite{shaban2017one,zhang2020sg,zhang2019canet,wangfew}. 
In contrast to these image analysis problems,
here we focus on the more challenging TAL problem.
Note that the model in \cite{hu2019silco}, though developed for object detection in images, can also work in the FS-TAL setting with trimmed support set.
More specifically, unlike our query adaptive Transformer for classifier adaptation at the sample level, it leverages self-attention to contrast the regional features 
exhaustively across the query and support samples.
We will compare with \cite{hu2019silco}
in our experiments (Table \ref{Tab1}).

\noindent{\bf Few-shot Temporal Action Localization}
FSL has been introduced to temporal action location recently \cite{yang2018one,zhang2020metal,yang2020localizing}.
Yang et al.~\cite{yang2018one}
propose the first FS-TAL setting with trimmed support set. It incorporates the sliding window idea in the matching network \cite{vinyals2016matching} to localize action instances in untrimmed query videos.
Later on, Zhang et al.~\cite{zhang2020metal} consider weak video-level annotation of untrimmed training videos.
Similar to our proposed setting, the latest work \cite{yang2018one} also focuses on a singe new class at one time. However, a common limitation with these existing FS-TAL problem settings stems from the assumption of trimmed support set.
As explained earlier,  trimmed videos do not exist naturally and need to be obtained with the same amount of manual annotation as our setting. Importantly, the ignorance of background content in the original untrimmed video leads to the failure to exploit useful context information.
 We will compare the two FS-TAL settings in our experiments (Table \ref{Tab1}).

\vspace{-0.15in}
\section{Proposed Methodology}

{\bf Problem Formulation} 
Given only a few videos from any unseen action class, we aim to learn a TAL model for that class.
%
For FS-TAL, we assume a base category set $C_{base}$ for training, and a novel category set $C_{novel}$ for testing.
For testing cross-class generalization,
we ensure that the two class sets are disjoint: $C_{base} \bigcap C_{novel} = \varnothing$.
Accordingly, the base and novel datasets are denoted as 
$D_{base} = \left \{ \left ( V_{i}, Y_{i} \right ), Y_{i} \in C_{base}\right \}$ and $D_{novel} = \left \{ \left ( V_{i}, Y_{i} \right ) , Y_{i} \in C_{novel}\right \}$ respectively.
Under the proposed new setting, each training video $V_{i}$ is associated with
segment-level annotation 
$Y_{i} = \left \{ (s_{t},e_{t},c), t \in \{1,..,M\}, c \in \textit{C}\right \}$ including $M$ segment labels each with the start and end time locations and action class. 
%
 In evaluation, 
for each task, we randomly sample a class $\textit{L} \sim C_{novel} $ from which \textit{K} and one labeled videos are randomly sampled to construct the support set \textit{S} and the query set \textit{Q} respectively. 
The labels of $S$ are accessible for model few-shot learning
whilst that of $Q$ are used for performance evaluation.

\subsection{Model Architecture}
\begin{figure*}[]
\begin{center}
  \includegraphics[height=5.8cm,width=\linewidth]{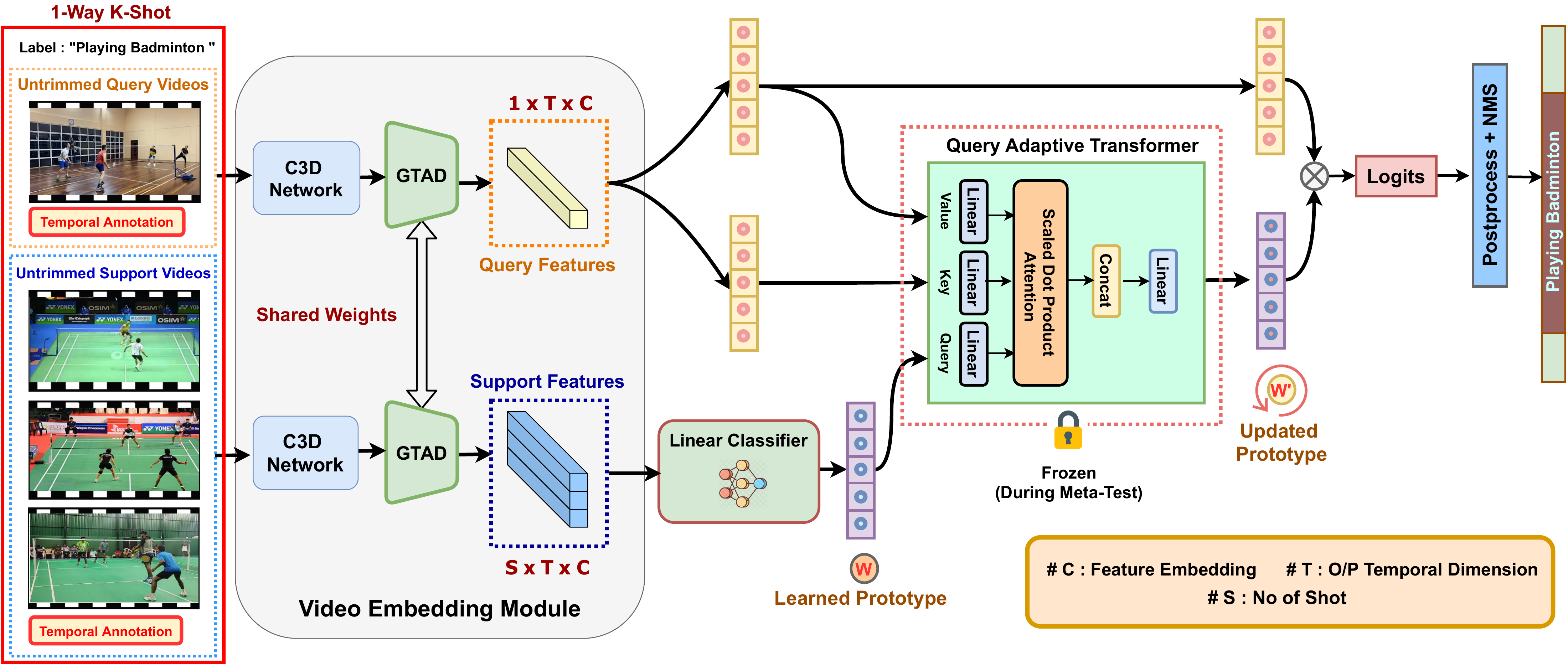}
\end{center}
\caption{\textbf{Overview of the proposed FS-TAL deep learning architecture.} 
There are two main modules:
(1) Video embedding for feature representation:
It is pre-trained on the whole training set, and shared by all different tasks 
for more effective knowledge transfer from training classes to test classes.
(2) Snippet classification for foreground prediction:
It is learned specifically for every individual task in two steps.
Initialized with the average of foreground snippet features, the first step learns the classifier on the support videos
in a supervised manner.
The second step further adapts the classifier weights
to every query video with a query adaptive Transformer.
The Transformer is meta-trained.
The final localization result is obtained by thresholding snippet-level classification scores and temporal non-maximum suppression.}
\vspace{-.2in}
\label{fig:network}
\end{figure*}

Our FS-TAL model is illustrated in Figure \ref{fig:network}. 
It consists of a 
task-generic video embedding module (Sec.~\ref{sec:video_embedding}),
and a task-specific snippet classification module (Sec.~\ref{sec:classification}).
We aim to achieve optimal video embedding and classification
for any new task with only a few (1 or 5) labeled support videos.
To that end, we share video embedding component across
all tasks, and exploit the classification component 
for tackling the task specificity.
With the output of task adapted classification on every snippet of a test video, we apply a non-parametric localization process
to obtain the segment predictions (Sec.~\ref{sec:inference}).



\subsection{Task-Generic Video Embedding}
\label{sec:video_embedding}

To capture action location information of a video, we construct a video embedding module with two components including feature backbone and snippet embedding.

\noindent\textbf{Feature backbone } 
In general, any video action models can be used such as 
C3D \cite{tran2015learning}, I3D \cite{carreira2017quo} and TSM \cite{lin2019tsm}.
For fair comparison with \cite{yang2020localizing},
we adopt the same backbone C3D as our default choice.
It is characterized by conducting 3D convolution and pooling operations in 2D spatial and 1D temporal dimensions simultaneously, capturing both appearance and motion information.  
%
Given an input video $V$, 
we extract RGB $X_{r} \in \mathbb{R}^{T \times d_1}$ 
and 
optical flow $X_{o} \in \mathbb{R}^{T \times d_1}$
features at the snippet level,
where $T$ denotes the number of snippets and $d_1$ denotes the feature dimension.
Each snippet is a short sequence of (\eg, 16 in our case) consecutive frames. 
We denote the concatenated features as $X = [X_r; X_o] \in \mathbb{R}^{T \times 2d_1}$.
As in most TAL methods \cite{xu2020gtad,xu2017r,lin2019bmn},
the feature backbone is pre-trained on a large video classification
dataset (\eg, Kinetics \cite{kay2017kinetics}) and 
then frozen to serve as a feature extractor.

\noindent\textbf{Snippet embedding }
Whilst C3D features have already encoded local motion information
due to using 3D convolution and optical flow,
long-term structural information is lacking but
critical for action localization.
To address this issue, we adopt an off-the-shelf temporal proposal model called GTAD \cite{xu2020gtad}. Other proposal models can be similarly integrated \cite{lin2018bsn,lin2019bmn,xu2020gtad}.
In particular, GTAD exploits a temporal graph and a semantic graph
for modeling long-term temporal and contextual information concurrently.
In our context, we utilize GTAD as a means for refining the C3D snippet features in a way that they become more sensitive to foreground (action content) and background.
We use the output of some intermediate layer of GTAD
as the snippet embedding.
The layer selection will be evaluated in our experiments (Sec. \ref{sec:experiments}).

Formally, taking C3D features $X \in \mathbb{R}^{T \times 2d_1}$ of a video as input, 
GTAD can output the snippet embedding as ${X}_{se} \in \mathbb{R}^{T \times C}$
where $C$ is the embedding dimension.
Support and query videos share the same GTAD model. {We denote $X^{s}_{se}$ and $X^{q}_{se}$ as the embedding of the support and query videos.}
Consider that snippet embedding would be largely shareable
among different tasks,
we train the GTAD model on the base dataset
in a standard supervised learning way.
The objective function includes a classification loss 
and a localization loss with respect to the ground-truth foreground mask  \cite{xu2020gtad}.
The trained GTAD and C3D form the video embedding module
which provides generic video representations for the
subsequent few-shot learning stage.

\subsection{Task-Specific Snippet Classification}
\label{sec:classification}
In our architecture, few-shot learning is focused on the snippet classification component for capturing each task's specificity.
We aim to build a binary classifier $h_{\phi}$ (with $\phi$ the parameters) that can distinguish foreground action from background content in a video.
Formally, the classifier model for predicting the foreground likelihood is a simple linear classifier as
\begin{equation}\label{eqn_1}
\textit{p}(t) = h_{\phi}(X_{se}(t)) = \sigma(\tau \cdot [cos(X_{se}(t), \phi)]),
\end{equation}
where $\sigma$ specifies the sigmoid function,
$\tau$ is a temperature hyper-parameter, and $cos$ is the cosine similarity. 
The snippet is indexed by $t \in \{1, \cdots, T\}$.

To make a classifier discriminative for each specific task, we introduce a two-step learning-and-adapting strategy.
In the first step,
we learn the classifier weights on the support set
in a supervised way.
In the second step,
we further adapt the support-set trained classifier weight
to every query video with a query adaptive Transformer model.
This aims to solve the intra-class variation problem.


\noindent{\bf New class adaptation} 
As the support set is composed of untrimmed videos with segment-level annotation, 
we can adapt the classifier to a new class with standard supervised learning.
Given the ground-truth annotation,
we label each snippet with foreground or background.
%
To train the classifier, we use the cross entropy loss as the objective function:
\begin{equation}\label{eqn_2}
\centering
\begin{aligned}
{L}_{ce} = - \frac{1}{2K}\sum_{k = 1}^{K} [{L}_{fg}\left ( {X}^s_{k}\right ) + {L}_{bg}\left ( {X}^s_{k}\right )], 
\end{aligned}
\end{equation}
\begin{equation}\label{eqn_3}
\centering
\begin{aligned}
{L}_{fg}\left ({X}_{k}^s\right ) = \frac{l_{fg} + l_{bg}}{\epsilon + l_{fg}}\sum_{t \in \{1,\cdots,T\}}
\hat{y}^{s}_k(t) \log[p_{k}^s(t)],
\end{aligned}
\end{equation}
\begin{equation}\label{eqn_4}
\centering
\begin{aligned}
{L}_{bg}\left ({X}_{k}^s\right ) = \frac{l_{fg} + l_{bg}}{\epsilon + l_{bg}}\sum_{t \in \{1,\cdots,T\}}
(1-\hat{y}^{s}_k(t)) \log[1-p_{k}^s(t)], 
\end{aligned}
\end{equation}
where $p_{k}^s(t)$
is the prediction of the $t$-th snippet $X_k^s(t)$
from the $k$-th support video.
$\epsilon$ is used to tackle extreme cases
such as zero background/foreground.
To balance the effect of foreground and background snippets in training, we introduce a balancing policy based on their sizes
($l_{fg}$ and $l_{bg}$).
The idea is intuitive -- less is more important.


The classifier can be trained by a small number of (\eg, 50$\sim$100) iterations.
We denote $\phi^*$ as the support-set trained classifier's weights.
Given only a handful of labeled samples,
how to initialize the classifier weights becomes more critical. Instead of random initialization, we found that the mean of foreground snippet's embedding serves as a stronger choice.



\paragraph{Query video adaptation}
Under the few-shot setting, a key challenge to overcome is the insufficient training samples in the support set for  capturing the intra-class invariance of the new class. As a result, training the classifier only on the support videos often fails to capture the discriminative informative generalizable to individual query videos.
To address this limitation, we propose a query adaptive Transformer model (with the parameters $\psi$) 
which is based on self-attention  \cite{vaswani2017attention}.



Taking an input in a triplet of $\{(query, key, value)\}$, our Transformer outputs an undated {\em query} with attentive association with respect to the {\em value}.
As our objective is to associate the classifier weights
$\phi^*$ with the query video $X_{se}^q$,
we set ${(query, key, value)} = (\phi^*, X_{se}^q, X_{se}^q)$.
The attentive learning is then formulated as
\begin{equation}\label{eqn_4}
A_{i}(\phi^*) = \phi^* + softmax(\frac{\phi^* W_{Q} ({X}_{se}^q W_{K})^{T}}{\sqrt{d}}) ({X}_{se}^q W_{V}),
\end{equation}
where $W_{Q}/W_{K}/W_{V}$ are learnable parameters (each is realized by a fully-connected layer) that projects 
the respective input to a $d$-dimension latent space.
In a multi-head attention (MA) design, we combine a set of independent $A_i$
to form a richer learning process:
\begin{equation}\label{eqn_5}
\phi^{**} = \underset{MA}{\underbrace{[A_{1}(\phi^*).. A_{m}(\phi^*)]}} + MLP(\phi^*) \ \in \ \mathbb{R}^{L \times 256}.
\end{equation}
The MLP block has one fully-connected layer with residual skip connection. Layer norm is applied before both the MA and MLP block.


{\em Learning objective }
After the classifier has been learned on both support and query videos, it can be applied to the query video.
We classify each snippet with Eq.~\eqref{eqn_1} with the foreground probability as:
\begin{equation}\label{eqn_6}
p{'}(t) : = h_{\phi^{**}}(X_{se}^q(t)).
\end{equation}
For training our Transformer ($\psi$), this prediction is then used to compute a cross-entropy loss (Eq.~\eqref{eqn_2}) as objective.
In meta-training, we conduct loss gradient back-propagation only once for each episode.
We denote $\psi^*$ as the optimized Transformer's parameters.

\subsection{Model Inference}
\label{sec:inference}
During testing, each time we are given a new task with one random unseen action class sampled from the novel dataset $D_{novel}$.
With the frozen video embedding module, we need to 
obtain a task-specific snippet classifier 
in two steps:
supervised learning with $K$ shots of support videos (Eq. \eqref{eqn_2}) 
and 
classifier weight adaptation on a query video 
by applying the meta-trained Transformer 
(note that our Transformer itself is frozen here).
The classifier is then applied to predict the foreground probability
of every snippet of a query/test video.

\noindent{\bf Action instance generation } 
After we obtain the snippet-level classification results,
we threshold on their {foreground probabilities}
and take those consecutive snippets as action instance candidates. 
To indicate the prediction confidence of each candidate,
we use the highest snippet foreground probability.
%
We then adjust the confidence scores using temporal soft Non-Maximal Suppression (NMS) \cite{bodla2017soft,lin2019bmn}. 
Finally, we select top $N$ candidates as the localization result.
\vspace{-0.3in}
\section{Experiments}
\label{sec:experiments}

{\bf Datasets}
We evaluate on two large-scale temporal action localization datasets. \textbf{ActivityNet-v1.3} \cite{caba2015activitynet} is a popular TAL benchmark.
It contains 19,994 temporally annotated untrimmed videos in 200 action categories. 
%
\textbf{THUMOS'14} \cite{jiang2014thumos} is another
widely used benchmark for action recognition and localization. There are 413 untrimmed videos from 20 different categories. 
The 20 classes are a subset of the 101 classes in UCF101 \cite{soomro2012ucf101}. 


\noindent{\bf Few-shot learning setting }
To facilitate performance comparison, we use the same class split as introduced in  \cite{yang2020localizing}.
For both datasets, we split the videos into single instance and multi-instance according to the number of action instances per video. 
For the single instance case, we divide the videos with multiple action instances into independent single-instance videos. Every newly generated video is no longer than 768 frames. 
For each of the two cases, 
we divide all the classes into three non-overlapping subsets for training (80\%), validation (10\%) and testing (10\%), respectively. 
The validation set is used for model parameter tuning and 
best model selection. We consider 1-shot and 5-shot.
In our setting we adopt untrimmed support set, 
as opposed to \cite{yang2020localizing} using trimmed videos. For each test, we use 5000 random tasks and report their average result.

\noindent{\bf Implementation Details} 
\noindent For each untrimmed video, we extract its RGB frames at 16 FPS and at the resolution of 256 × 256. We averagely divide each video into 100 (256 for THUMOS) non-overlapping snippets and sample 8 frames for each snippet (\ie, $T=100$). As \cite{yang2020localizing}, we filter out videos having less than 768 frames.
We consider single-instance and multi-instance test videos, separately.
The dimension of C3D feature is 500 (\ie, $d_1=500$). 
We take the penultimate layer's output (Layer-5) of GTAD's localization module as video embedding (256-D).
The latent feature dimension $d$ (Eq.~\eqref{eqn_4}) of our query adaptive Transformer is 256.
Dropout is used in our Transformer to alleviate model overfitting.
We set the NMS threshold of 0.7/0.6 for ActivityNet/THUMOS.
As the final TAL result, we take top 100/200 for ActivityNet/THUMOS. 
We adopt the Adam optimizer \cite{sutskever2013importance} with learning rate 0.004.
We train the model for $50$ epochs each with $200$ episodes.




\begin{table*}[!hbtp]
\resizebox{1\columnwidth}{!}{
\centering
\Large
\begin{tabular}{@{}c|ccccc|c|ccccc|c||ccccc|c|ccccc|c@{}}
\toprule

\multicolumn{13}{c||}{\textbf{Single instance videos}}                 & \multicolumn{12}{c}{\textbf{Multi-instance videos}} \\ \midrule
         & \multicolumn{6}{c|}{ActivityNet-v1.3} & \multicolumn{6}{c||}{THUMOS'14}       & \multicolumn{6}{c|}{ActivityNet-v1.3}  & \multicolumn{6}{c}{THUMOS'14}       \\ \midrule
map@     & 0.5  & 0.6  & 0.7 & 0.8 & 0.9 & mean & 0.5 & 0.6 & 0.7 & 0.8  & 0.9  & mean & 0.5  & 0.6  & 0.7  & 0.8 & 0.9 & mean & 0.5 & 0.6 & 0.7 & 0.8  & 0.9  & mean \\ \midrule
\multicolumn{13}{c||}{\textit{1 Shot}}                                                  & \multicolumn{12}{c}{\textit{1 shot}}                                        \\ \midrule
Hu et al. \cite{hu2019silco} & 41.0 & 33.0 & 27.1 & 15.9 & 6.8  & 24.8 & - & - & - & - & - & -  & 29.6 & 23.2 & 12.7 & 7.4 & 3.1  & 15.2 & - & - & - & - & - & - \\
Feng et al. \cite{feng2018video} & 43.5 & 35.1 &  27.3 & 16.2 & 6.5 & 25.7  & - & - & - & - & - & - & 31.4 & 25.5 & 16.1 & 8.9 & 3.2 & 17.0  & - & - & - & - & - & - \\
Yang et al. \cite{yang2020localizing} & 53.1 & 40.9 & 29.8 & 18.2 & 8.4  & 29.5 & 48.7 & - & - & - & - & -  & 42.1 & 36.0 & 18.5 & 11.1 & 7.0  & 22.9 & - & - & - & - & - & - \\ \midrule
Ours & 55.1 & \textbf{45.2} & 35.5 & \textbf{25.3} & \textbf{13.2} & \textbf{32.5}  & 49.2 & 36.9 & \textbf{24.3} & \textbf{16.5} & \textbf{10.1} & \textbf{27.2} & 44.1 & 37.8 & \textbf{29.5} & 21.4 & \textbf{11.5} & 25.8  & 7.3 & 4.2 & 3.1 & 2.0 & 1.5 & 3.7 \\
Ours$^\dagger$ & \textbf{55.6} & 44.6 & \textbf{35.7} & 24.6 & 12.7 & 31.8  & \textbf{51.2} & \textbf{38.1} & 22.7 & 14.8 & 9.2 & 27.0 & \textbf{44.9} & \textbf{38.0} & 29.2 & \textbf{21.4} & 11.2 & \textbf{25.9}  & \textbf{9.1} & \textbf{6.8} & \textbf{4.9} & \textbf{3.5} & \textbf{2.3} & \textbf{5.3} \\ \midrule
\multicolumn{13}{c||}{\textit{5 Shot}}                                                  & \multicolumn{12}{c}{\textit{5 shot}}                                        \\ \midrule
Buch et al. \cite{buch2017sst} & 39.7 & 33.6 & 27.0 &  14.0 & 4.6 & 23.3  & 35.7 & 29.4 & 20.8 & 11.7 & 3.4 & 20.2 & 30.4 & 25.1 & 19.6 & 12.9 & 6.6 & 18.9  & 2.7 & 1.9 & 1.4 & 0.9 & 0.4 & 1.5 \\
Hu et al. \cite{hu2019silco} & 45.4 & 35.0 & 29.9 & 17.6 & 5.2 & 27.0  & 42.2 & 32.6 & 20.3 & 13.7 & 5.2 & 22.8 & 38.9 & 27.2 & 18.3 & 12.7 & 7.3 & 20.9  & 6.8 & 3.1 & 2.2 & 1.8 & 1.3 & 3.1 \\
Yang et al. \cite{yang2020localizing} & 56.5 & 47.0 &  37.4 &  21.5 & 11.9 & 34.9  & 51.9 & 42.7 & 24.4 & 17.7 & 10.1 & 29.3 & 43.9 & 37.4 & 20.2 & 13.4 & 7.7 & 24.5  & 8.6 & 5.6 & 3.8 & 2.5 & 1.7 & 4.4 \\ \midrule
Ours & 63.0 & \textbf{54.5} & \textbf{44.2} & 30.9 & 15.8 & 38.4  & 54.3 & 43.6 & \textbf{35.8} & \textbf{24.5} & 12.2 & 31.6 & 48.2 & 39.1 & 29.7 & 22.5 & \textbf{12.8} & 28.2  & 10.4 & 7.1 & 5.7 & 4.8 & 2.9 & 5.4 \\ 
Ours$^\dagger$ & \textbf{63.8} & 54.2 & 43.9 & \textbf{31.4} & \textbf{16.4} & \textbf{38.5}  & \textbf{56.1} & \textbf{47.2} & 32.4 & 24.3 & \textbf{13.7} & \textbf{32.7} & \textbf{51.8} & \textbf{42.7} & \textbf{32.6} & \textbf{23.4} & 11.9 & \textbf{30.2}  & \textbf{13.8} & \textbf{11.3} & \textbf{8.4} & \textbf{6.3} & \textbf{4.2} & \textbf{7.1} \\ \bottomrule
\end{tabular}
}
\\\\
\caption{
FS-TAL results (\%). 
$^\dagger$: Using untrimmed support set (i.e., the new setting).
}
\label{Tab1}
\vspace{-0.8cm}
\end{table*}

\subsection{Comparison with state-of-the-art} 
{\bf Competitors }
For comparative evaluation, we consider a few-shot object detection model \cite{hu2019silco}, 
a one-shot video re-localization model \cite{feng2018video},
and the latest FS-TAL model \cite{yang2020localizing}.
Because \cite{feng2018video} cannot tackle multiple support videos, we compare with a modified version of temporal action proposal model SST \cite{buch2017sst} for 5-shot case. 
As in \cite{yang2020localizing}, a fusion layer is added on top of SST's GRU layer to incorporate the support video features, and the proposal with the largest confidence score is taken as the prediction.
All the methods use the same C3D video feature backbone.
For all the competitors,
we use trimmed support set to keep their original designs.
We evaluate the proposed model under both the previous setting (trimmed support set) and our new setting (untrimmed support set). 
This allows for absolute fair model comparison as well as
setting comparison.
When feeding trimmed support videos into our model, 
the background loss term $L_{bg}$ in Eq.~\eqref{eqn_2} will become zero; without any other formulation change, our model can be applied to the old setting. The difference is that now the Transformer is used to adapt a foreground template/prototype to each query video, instead of a foreground/background classifier. Note that none of the existing methods can be easily extended to operate under our new setting.


\noindent{\bf Results } 
%
%
The results are compared in Table \ref{Tab1}.
It is evident that our method achieves the best performance in all test settings when using the same trimmed support set. 
This suggests the superiority of our model over all alternative designs, verifying the proposed few-shot learning architecture.
%
The margin is even larger in more strict metrics.
Importantly, we see that the margin further 
increases in 5-shot case, indicating the superior capability of our method in leveraging multiple training videos.
This is mainly due to the proposed query adaptive Transformer that can amplify the benefit of larger support-set
via attentive query video adaptation,
 which is lacking in all existing methods.
%
In the multi-instance setting on THUMOS'14,
all the methods do not work well due to longer videos
and short action instances.
However, it is still encouraging that our model can double or triple the performance of alternatives at mAP@0.6-0.9
in such challenging test.

We further examine the two FS-TAL problem settings 
with the proposed method. 
We make the following observations.
In the single-instance setting,
the model performance is marginally better 
in previous setting with trimmed videos in most cases.
\xz{Our observation suggests that this is potentially due to lack of background diversity.}
However, when it comes to the more practical and  challenging multi-instance setting, the opposite is true especially in the 5-shot case.
This indicates that background helps model learning
with useful context cues co-existing with action instances.
Given these observations, 
we consider that the proposed setting is not only more practical
but also provides more information for better modeling,
as compared to the previous settings.

\vspace{-0.15in}
\subsection{Effect of Query Video Adaptation}
\begin{figure*}[]
\begin{center}
  \includegraphics[scale=0.42]{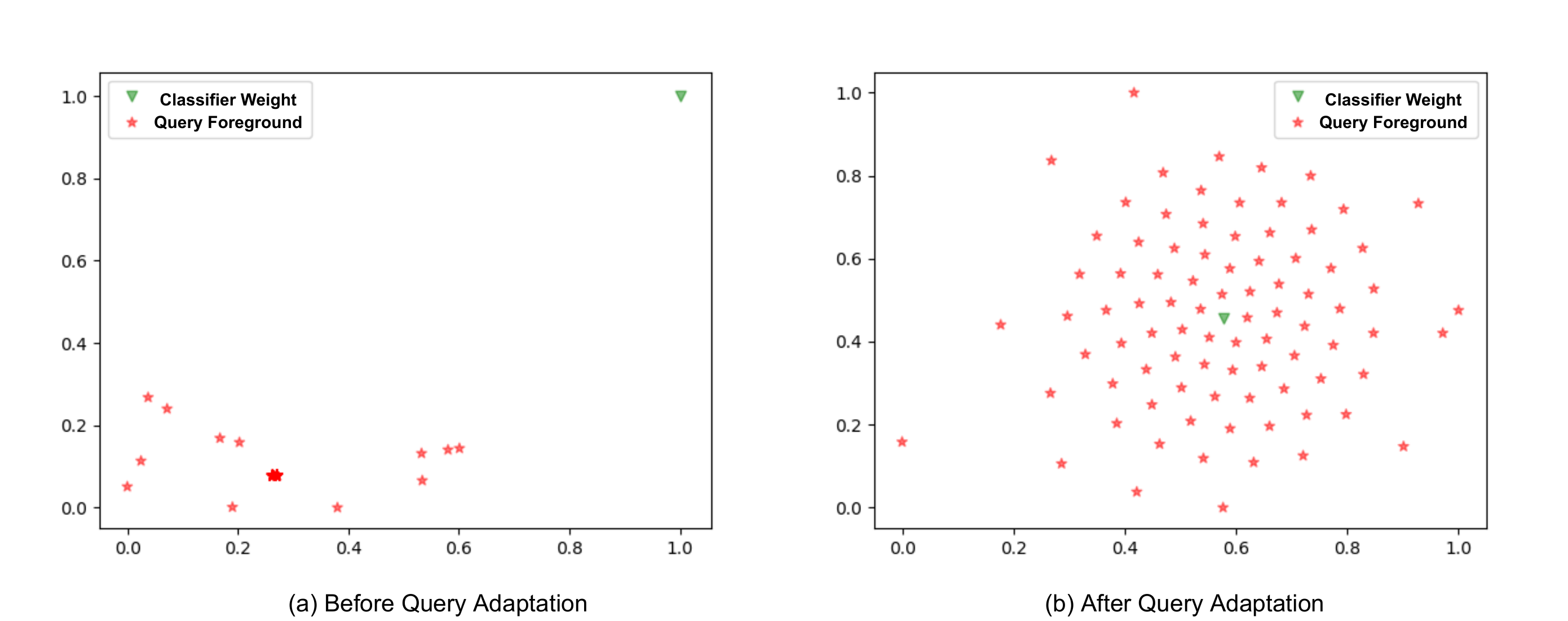}
\end{center}
\caption{\textbf{The effect of Query Video Adaptation with t-SNE visualization.} 
It is shown that with the proposed query video adaptation,
the classifier weight can be effectively pushed to be aligned with the foreground content of the query video sample.
This improves learning the intra-class invariance of the new class.
}
\vspace{-0.2in}
\label{fig:tsne}
\end{figure*}

In Section \ref{sec:classification}
we introduce a query adaptive Transformer
for fast adapting the support-set trained classifier's weights to each query video.
This aims to solve intra-class variation with FS-TAL 
as there is no sufficient training samples in support set to capture
this variation.
There may exist big appearance difference between the support and query video action instances
(see Figure 2 in Supplementary). 
Query video adaptation is thus critical.
From Table \ref{tab:qva} we can see that without the proposed query video the performance will drop significantly ($3\sim 8$\%) in 1/5-shot settings of both datasets. 
This verifies the importance of learning the intra-class invariance problem and the ability of our Transfer model in adapting the classifier's weight to each query video, \ie, video-specific adaptation.
\textcolor{black}{In Figure \ref{fig:tsne} we visually show that our query adaptive transformer is effective in adapting the classifier's weight to capture the specificity of the query video's foreground content.}

\begin{table}[!hbt]
\centering
\resizebox{0.7\columnwidth}{!}{
\begin{tabular}{@{}ccccc@{}}
\toprule
\multicolumn{1}{c|}{Dataset}  & \multicolumn{2}{c|}{ActivityNet} & \multicolumn{2}{c}{Thumos} \\ \midrule
\multicolumn{1}{c|}{mAP}      & 0.5  & \multicolumn{1}{c|}{mean} & 0.5          & mean         \\ \midrule
\multicolumn{5}{c}{\textbf{Without QVA}}                                                  \\ \midrule
\multicolumn{1}{c|}{Ours @ 1-shot} &  37.3    & \multicolumn{1}{c|}{21.7}     &  3.6         &      2.3        \\
\multicolumn{1}{c|}{Ours @ 5-shot} &  43.8    & \multicolumn{1}{c|}{25.3}     &   7.9          &    4.0          \\ \midrule
\multicolumn{5}{c}{\textbf{With QVA}}                                               \\ \midrule
\multicolumn{1}{c|}{Ours @ 1-shot} &   44.9 ($\uparrow \textbf{7.6}$)   & \multicolumn{1}{c|}{25.9 ($\uparrow \textbf{4.2}$)}   &   9.1 ($\uparrow \textbf{5.5}$)          &      5.3  ($\uparrow \textbf{3.0}$)      \\
\multicolumn{1}{c|}{Ours @ 5-shot} &  51.8 ($\uparrow \textbf{8.0}$)    & \multicolumn{1}{c|}{30.2 ($\uparrow \textbf{4.9}$)}     &   13.8  ($\uparrow \textbf{5.9}$)         &     7.1  ($\uparrow \textbf{3.1}$)      \\ \bottomrule
\end{tabular}
}
\vspace{0.15in}
\caption{\textbf{Effect of query video adaptation (QVA)} 
in the multi-instance setting.
}
\label{tab:qva}
\vspace{-0.2cm}
\end{table}

\begin{figure}[h]
\begin{center}
  \includegraphics[scale=0.12]{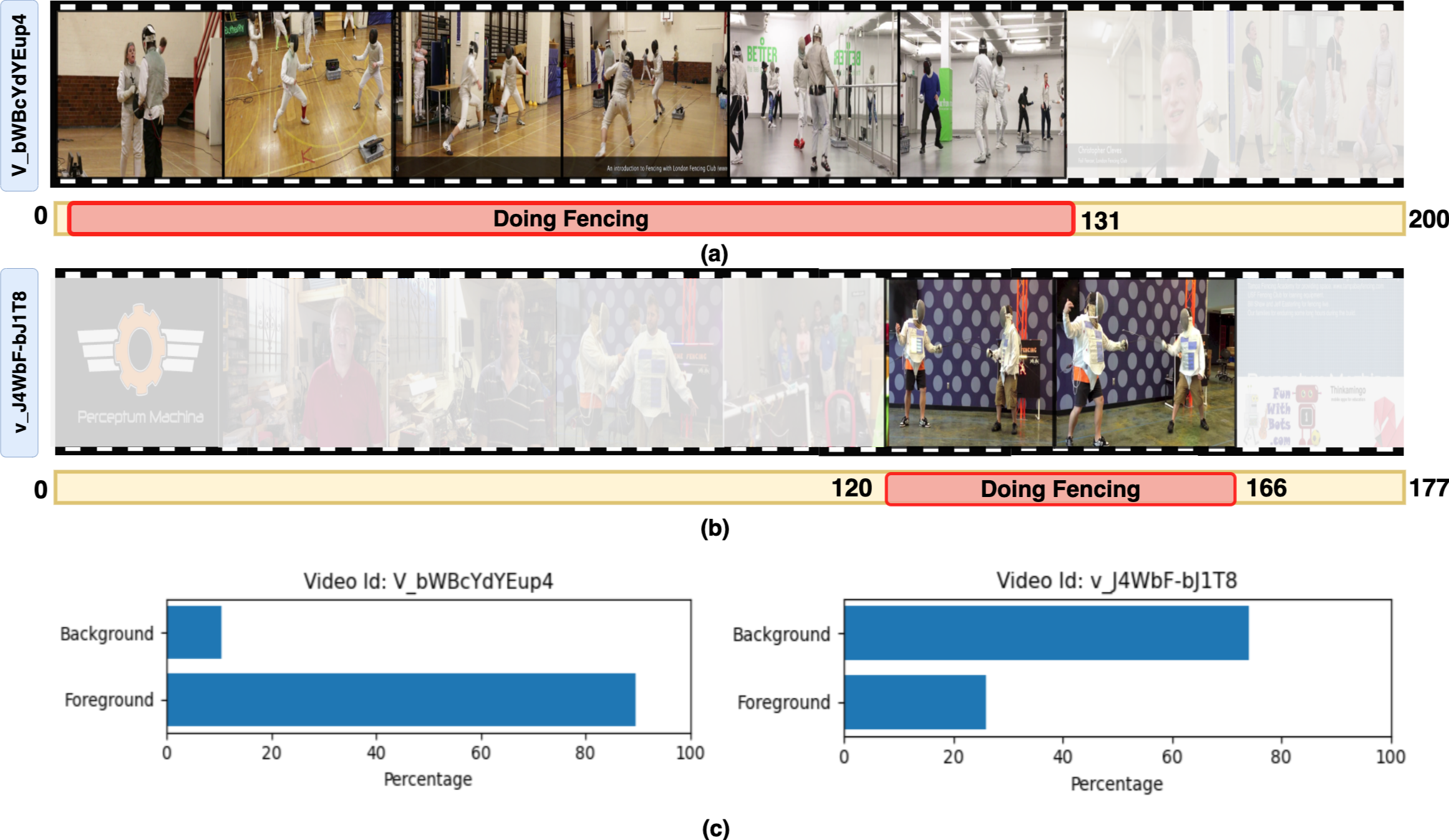}
\end{center}
\caption{\textbf{Intra-class variation} example in 
the ``Doing Fencing'' class
on ActivityNet-v1.3. 
As can be seen, the two videos present clear difference
in viewpoint, scene setup, background, illumination,
as well as instance length (c).
}
\label{fig:intra_class}
\end{figure}
\vspace{-0.2in}
\subsection{Cross-Domain Localization} 
Following the above single domain (dataset) FS-TAL evaluation,
we further introduce a more challenging and more realistic cross-domain setting. 
As THUMOS'14 and ActivityNet-v1.3 present large differences in action instance length and background characteristics, they are suitable for cross-domain evaluation.
We consider the single-instance setting.
We compare our method with the state-of-the-art model \cite{yang2020localizing}.



\noindent {\bf $\textbf{THUMOS} \rightarrow \textbf{ActivityNet}$ }
In the first cross-domain experiment, we train a model on the base classes of THUMOS'14 (source domain) and test the model on the novel classes of ActivityNet-v1.3 (target domain). 
The results are reported in Table \ref{tab:cross_domain}.
It is shown that the performance advantage of our method 
remains compared to the single-domain setting.
For example, the mAP@0.5 margin of our model over \cite{yang2020localizing} is 4.8\%/6.2\%
in the 1/5-shot cases.
%
Comparing the single-domain results in Table \ref{Tab1},
we can see that domain shift indeed negatively affects
the performance of both models.


\noindent{\bf $\textbf{ActivityNet} \rightarrow \textbf{THUMOS}$ } 
The second experiment considers the opposite transfer direction. At large we have similar observations with
our model again outperforming \cite{yang2020localizing} in both 1/5-shot setting.
This suggests that our model can generalize to different transfer setups with consistent performance advantages.

\begin{table}[]
\centering
\resizebox{0.7\columnwidth}{!}
{
\begin{tabular}{@{}c|cc|cc@{}}
\toprule
Cross Domain & \multicolumn{2}{c|}{$\textbf{Thumos} \rightarrow \textbf{ActivityNet}$} & \multicolumn{2}{c}{$\textbf{ActivityNet} \rightarrow \textbf{Thumos}$} \\ \midrule
mAP          & 0.5                      & mean                     & 0.5                      & mean                     \\ \midrule
Yang et al. \cite{yang2020localizing} @ 1-shot & 41.1                      & 25.2                      & 36.2                      & 21.4                      \\
Ours @ 1-shot     & $\textbf{45.9}$                      & $\textbf{26.6}$                      & $\textbf{38.1}$                      & $\textbf{22.5}$                      \\ \midrule \midrule
Yang et al. \cite{yang2020localizing} @ 5-shot & 48.2                      & 27.8                      & 37.5                      & 23.6                      \\
Ours @ 5-shot     & $\textbf{54.4}$                      & $\textbf{31.6}$                     & $\textbf{43.8}$                      & $\textbf{27.2}$                      \\ \bottomrule
\end{tabular}
}
\vspace{0.15in}
\caption{\textbf{Cross-domain FS-TAL.}
}
\label{tab:cross_domain}
\vspace{-0.5cm}
\end{table}

\subsection{Effect of Video Embedding Module}
We evaluate the generality of our FS-TAL architecture in different video embedding designs.
In this test we select BMN \cite{lin2019bmn}.
Table \ref{tab:embed} shows that BMN is slightly inferior to GTAD for video embedding, which is consistent with the previous finding \cite{xu2020gtad}.

\subsection{Inference Efficiency}
In inference, our model runs a small number of iterations
for learning the linear classifier's weights on the support set, which increases slightly
the computational overhead. 
We conduct a quantitative cost analysis 
in 5-shot multi-instance setting on ActivityNet-v1.3.
We compared to the state-of-the-art model \cite{yang2020localizing}.
For both methods, we track the speed of 100 FS-TAL tasks
on a machine with one RTX2080Ti GPU.
Table \ref{tab:speed} shows that
our method has very similar inference speed as \cite{yang2020localizing},
without efficiency disadvantage.

\begin{minipage}{0.95\textwidth}
\vspace{0.15in}
  \begin{minipage}[b]{0.42\columnwidth}
    \centering
    \small
    \begin{tabular}{@{}c|c|c@{}}
    \toprule
    Dataset  & \multicolumn{2}{c}{ActivityNet-v1.3} \\ \midrule
    mAP & 0.5         & mean        \\ \midrule
    BMN \cite{lin2019bmn}     & 61.6 & 37.5  \\
    GTAD \cite{xu2020gtad}    & 63.8 & 38.5  \\ \bottomrule
    \end{tabular}
    \vspace{0.15in}
    \captionof{table}{\textbf{Effect of video embedding} in the 
    5-shot multi-instance setting.
    }
    \label{tab:embed}
 \end{minipage}
  \hfill
 \begin{minipage}[b]{0.53\textwidth}
    \centering
    \small
    \begin{tabular}{@{}c|c@{}}
    \toprule
    Dataset  & {ActivityNet-v1.3} \\ \midrule
    Metrics & Speed (seconds / task)      \\ \midrule
    Yang et al. \cite{yang2020localizing}     
    & 0.81  \\
    Ours    
    & 0.83  \\ \bottomrule
    \end{tabular}
    \vspace{0.15in}
    \captionof{table}{\textbf{Inference efficiency test} in the 
    5-shot multi-instance setting with a RTX2080 GPU.
    }
    \label{tab:speed}
 \end{minipage}
\end{minipage}

\subsection{Qualitative Analysis}
\begin{figure*}[h]
\begin{center}
  \includegraphics[scale=0.44]{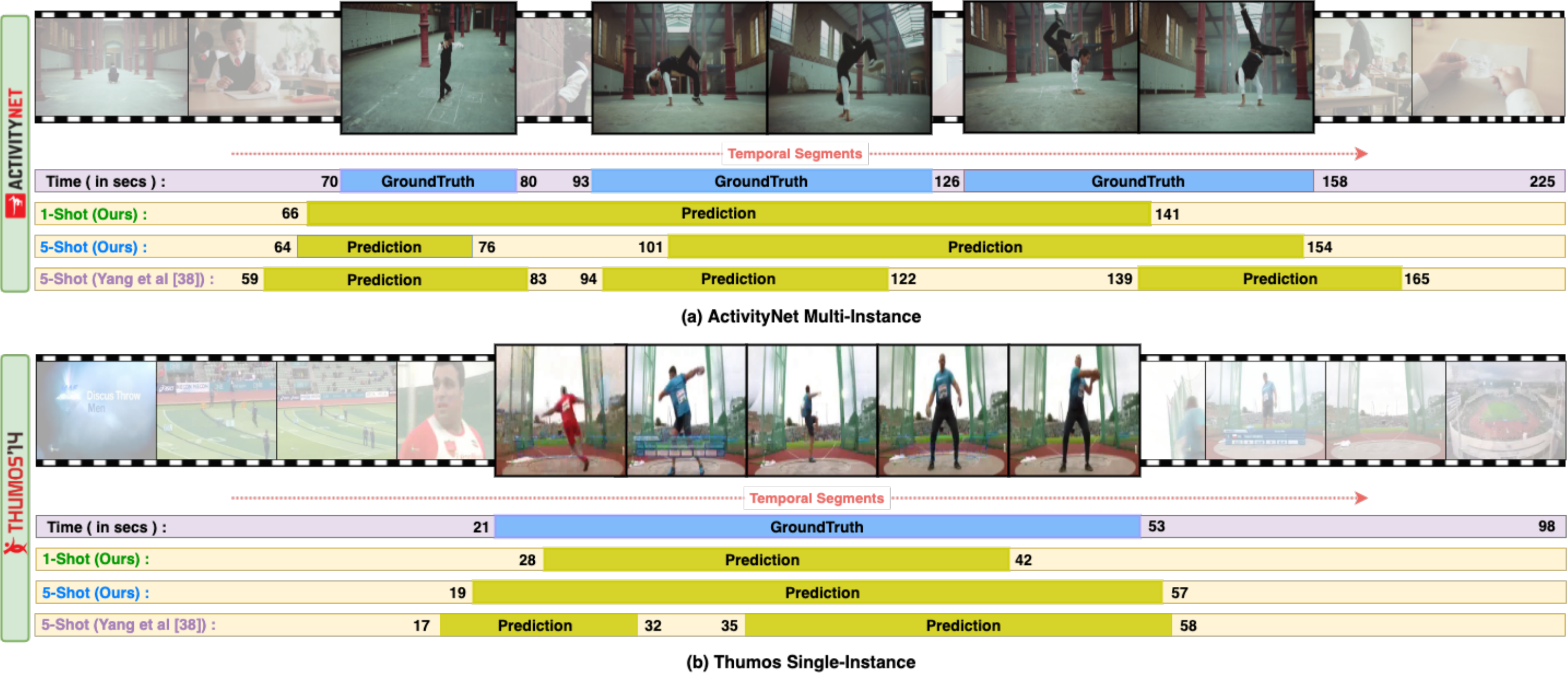}
\end{center}
\caption{\textbf{Qualitative results} of (a) ``BreakDancing`` class on ActivityNet and (b) ``Throw Discuss`` class on THUMOS. 
}
\vspace{-.15in}
\label{fig:example}
\end{figure*}

For visual analysis, we provide two qualitative examples in Figure \ref{fig:example}.
To visualize the intra-class variation challenge which our  method in particular the proposed query  adaptive  Transformer aims to address,
we show some common examples in 
Figure \ref{fig:intra_class}.

\subsection{Choice of Video Embedding Layer}
\begin{figure}[!htbp]
\begin{center}
  \includegraphics[scale=0.65]{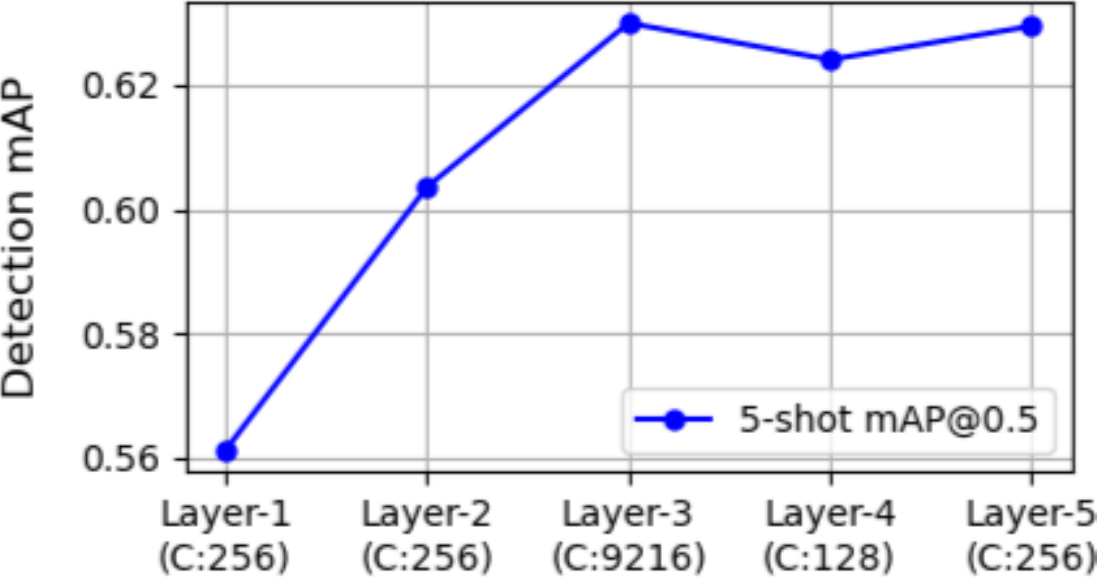}
\end{center}
\caption{\textbf{Ablation of GTAD video embedding layer} in the single-instance setting on ActivityNet-v1.3.
The number in round bracket is the embedding dimension.
}
\label{fig:gtad_layer}
\end{figure}

We examine which layer of GTAD \cite{xu2020gtad}
is a good choice for video snippet embedding.
In particular, we test five GTAD layers.
The result curve in Figure \ref{fig:gtad_layer} shows that
deeper layers are usually better than shallow ones,
suggesting that snippet-level contextual information is useful for action localization.
We select the layer-5 as our embedding layer 
as it has best cost-effectiveness.

\section{Conclusion}
We have presented a new and more practical few-shot temporal action localization (FS-TAL) problem.
Unlike all existing settings, 
in our setting a new action class is represented by untrimmed support set with useful background segments to provide contextual information.
We introduce a novel FS-TAL architecture
that effectively transfers class-generic representation
knowledge from training classes to any unseen test classes whilst adapting the model to any new class.
To solve the large intra-class variation problem,
we introduce a query adaptive Transformer
that further dynamically adapts the support-set trained classifier's weights to each query video.
Experiments on two popular TAL datasets
verify the superiority of our method over existing alternatives
in both the newly proposed setting with untrimmed labeled support set and previous settings with trimmed counterpart.
Moreover, our method remains to be advantageous
under a more realistic and challenging cross-domain setting.

\bibliography{egbib}
\end{document}